\documentclass{article}

\usepackage{amsmath}
\usepackage{graphicx} 
\usepackage{booktabs}
\usepackage{multirow}

 \usepackage[dblblindworkshop, final]{neurips_2026}
\workshoptitle{Foundations of Language Model Security}

\usepackage[utf8]{inputenc} 
\usepackage[T1]{fontenc}    
\usepackage{hyperref}       
\usepackage{url}            
\usepackage{booktabs}       
\usepackage{amsfonts}       
\usepackage{nicefrac}       
\usepackage{microtype}      
\usepackage{xcolor}         

\title{Depth-Dependent Safety Geometry in Vision-Language Models: Layer-Wise Localization and Bidirectional Refusal Control}

\author{%
  Muhammad Zeeshan Karamat \\
  Virginia Tech\\
  \texttt{mzeeshan@vt.edu} \\
  \And
  Syed Talal Hassan \\
  Virginia Tech\\
  \texttt{syedtalal@vt.edu} \\
  \And
  Christiana Chamon \\
  Virginia Tech\\
  \texttt{ccgarcia@vt.edu} \\
}

\begin{document}

\maketitle

\begin{abstract}
Vision-Language Models (VLMs) are widely deployed in multimodal systems, yet the internal mechanisms governing their safety behavior remain poorly understood. We introduce a lightweight inference-time framework for localizing and manipulating safety-sensitive representations without retraining or modifying model parameters. On Qwen3-VL-4B, activation divergence analysis reveals non-uniform safety geometry across depth. Causal interventions show that early vision layers support strong bidirectional refusal control through directional steering, with unsafe rejection spanning 6\%--96\%, whereas later layers are more responsive to activation magnitude scaling and exhibit a substantial safety--selectivity trade-off. We further observe qualitatively similar depth-dependent activation geometry in InternVL2.5-4B despite differences in precise layer location and magnitude. Together, these results distinguish representational separability from causal controllability, suggest that depth-dependent safety geometry recurs across VLM architectures, and expose a white-box representation-level attack surface that motivates inference-time stabilization of safety-sensitive representations.
\end{abstract}

\section{Introduction}

Vision-Language Models (VLMs) are increasingly used in multimodal systems such as visual question answering, image captioning, and interactive agents. As their deployment expands, understanding the mechanisms underlying safety behavior is critical to assessing robustness and trustworthiness. Although contemporary VLMs are aligned to refuse harmful requests, it remains unclear where safety-relevant information is represented, which internal components causally influence refusal, and how these mechanisms may fail under adversarial or inference-time perturbations.

Recent work suggests that safety alignment is not uniformly distributed across network depth. In VLMs, this raises the question of whether safety is governed primarily by downstream language semantics or whether safety-relevant structure already emerges within the visual representation pipeline. Existing approaches identify layer-wise differences in safety behavior, but it remains unclear whether safety-sensitive representations in frozen VLMs are controlled through activation magnitude, specific geometric directions, or different mechanisms at different depths. Moreover, representational separability alone does not imply causal control: a layer may distinguish safe from unsafe inputs without the corresponding direction directly governing refusal.

We introduce a lightweight inference-time framework for identifying and manipulating safety-sensitive regions in frozen VLMs without modifying model parameters. Our framework combines layer-wise activation divergence analysis with controlled representation-level interventions. We first characterize the geometric separation between safe and unsafe representations across the vision encoder and language model, then intervene on selected vision layers using activation magnitude scaling and Concept Activation Steering to test their causal influence on refusal behavior.

On Qwen3-VL-4B, we find a depth-dependent distinction between representational separability and causal controllability. Early vision layers permit strong bidirectional refusal control through directional steering, whereas later layers are comparatively less responsive to steering but more sensitive to activation magnitude scaling. Notably, deeper representations can remain geometrically separable even when steering the corresponding direction produces limited behavioral change. We further repeat the divergence analysis on InternVL2.5-4B and observe qualitatively similar depth-dependent geometry despite differences in precise layer location and magnitude, providing cross-model representational evidence that the observed structure is not unique to Qwen3-VL-4B.

Together, these findings reveal depth-dependent safety geometry whose causal influence varies across the vision encoder, enabling targeted refusal control while exposing a representation-level attack surface.

\noindent\textbf{Contributions.} Our main contributions are:
\begin{itemize}
    \item We introduce an inference-time framework for localizing and probing safety-sensitive representations in frozen VLMs.
    \item On Qwen3-VL-4B, we show that early vision layers support bidirectional refusal control, while later layers are more sensitive to magnitude scaling.
    \item We expose a representation-level refusal-bypass attack surface and observe qualitatively similar depth-dependent safety geometry in InternVL2.5-4B.
\end{itemize}

\section{Threat Model}

We consider a frozen Vision-Language Model (VLM) whose parameters cannot be modified at inference time. The adversary interacts with the model through multimodal inputs, such as adversarial images or prompts, with the goal of inducing internal representation shifts that weaken or bypass refusal behavior while preserving coherent outputs.

To analyze this threat in a controlled setting, we use forward hooks to directly perturb intermediate activations. These interventions are a white-box analysis tool rather than an input-level attack: they isolate specific layers and representation directions and measure their causal influence on refusal behavior. Thus, our experiments characterize a \emph{representation-level attack surface} and simulate worst-case internal shifts that adversarial inputs might induce, but do not establish that the same shifts are realizable solely through external inputs.

Our objective is to identify internal safety mechanisms whose instability could enable such attacks. Constructing an end-to-end input-level attack that reproduces these representation shifts is outside the scope of this work.

\section{Related Work}

\paragraph{Localization and Internal Mechanisms of Safety.}
Recent work suggests that safety alignment is not uniformly represented throughout large models. In LLMs,~\cite{li2025safetylayersalignedlarge} identify contiguous ``safety layers'' that disproportionately influence malicious-versus-benign behavior, while ~\cite{qi2025safetyalignment} show that safety alignment can remain shallow in the output distribution.~\cite{pan2025hiddendimensionsllmalignment} further identify multiple orthogonal safety-related directions associated with refusal and other behavioral properties. In multimodal models,~\cite{bachu2025layerwisealignmentexaminingsafety} show that harmful information is unevenly distributed across VLM image-encoder layers and introduce the Image enCoder Early-exiT (ICET) vulnerability, where bypassing selected layers increases harmful generation. Their Layer-Wise PPO defense improves alignment across intermediate representations. These studies establish that safety varies across depth, but do not determine whether safety-sensitive regions exert control through activation magnitude, geometric directions, or different mechanisms at different depths.

\paragraph{Representation Engineering and Activation Steering.}
Concept Activation Vectors~\cite{kim2018tcav} established that human-interpretable concepts can correspond to directions in activation space. Subsequent representation-engineering methods directly manipulate such internal structure to control model behavior.~\cite{zou2023representation} formulate representation engineering as a general framework for monitoring and control,~\cite{li2023iti} introduce Inference-Time Intervention to improve truthfulness through activation shifts, and Activation Addition~\cite{turner2023activation} uses contrastively derived steering vectors to modify behavior without optimization. In the safety domain,~\cite{arditi2024refusal} show that refusal in LLMs is mediated by a low-dimensional direction whose removal suppresses refusal and whose addition can induce refusal on benign inputs. Our work extends this perspective to VLM vision encoders and studies how directional controllability changes across depth.

\paragraph{VLM Safety and Inference-Time Defenses.}
VLM defenses span both training-time and inference-time approaches.~\cite{weng2025adversaryawaredpoenhancingsafety} propose Adversary-Aware DPO, which incorporates adversarial perturbations into preference optimization. VLM-Guard~\cite{liu2025vlmguardsafeguardingvisionlanguagemodels} instead operates at inference time using a safety direction derived from the underlying language model. Most closely related, ASTRA~\cite{wang2025astra} constructs harmful-response steering vectors from adversarial visual features and adaptively steers VLM activations away from harmful directions without modifying model weights.

Our work differs by using activation steering as a causal diagnostic in addition to an intervention. We localize safety-sensitive regions across depth, compare directional steering with magnitude scaling, and distinguish representational separability from causal controllability. On Qwen3-VL-4B, early vision layers permit strong bidirectional refusal control, whereas deeper representations are more responsive to magnitude perturbations despite remaining geometrically separable. InternVL2.5-4B further exhibits qualitatively similar depth-dependent activation geometry, providing cross-model representational evidence that this structure is not unique to a single architecture.

\section{Methodology}

Our framework consists of three stages: (1) layer-wise activation divergence analysis to identify candidate safety-sensitive regions, (2) activation scaling to measure sensitivity to representation magnitude, and (3) directional steering to test whether safety-related geometry causally influences refusal behavior.

\subsection{Activation Divergence Analysis}

We compare hidden representations produced by safe and unsafe inputs across the vision encoder and language model. We collect 100 safe examples
$\mathcal{S}=\{x_i\}$ and 100 unsafe examples
$\mathcal{U}=\{u_j\}$. For each layer $\ell$, we extract a pooled hidden representation $\mathbf{h}_\ell(x)$ and form
\begin{equation}
\mathcal{H}_\ell^{\mathcal{S}}=\{\mathbf{h}_\ell(x_i)\}, \qquad
\mathcal{H}_\ell^{\mathcal{U}}=\{\mathbf{h}_\ell(u_j)\}.
\end{equation}

At each layer, we construct Safe--Safe (S--S), Unsafe--Unsafe (U--U), and Safe--Unsafe (S--U) pairs. We randomly sample pairs from each category and repeat the procedure for $r$ runs to obtain stable layer-wise estimates.

For a pair $(a,b)$, cosine similarity is
\begin{equation}
\mathrm{cos}_\ell(a,b)=
\frac{\langle \mathbf{h}_\ell(a),\mathbf{h}_\ell(b)\rangle}
{\|\mathbf{h}_\ell(a)\|_2\|\mathbf{h}_\ell(b)\|_2},
\end{equation}
which we convert to angular distance in degrees:
\begin{equation}
\theta_\ell(a,b)=
\frac{180}{\pi}
\arccos\!\left(\mathrm{cos}_\ell(a,b)\right).
\end{equation}

We measure excess cross-condition angular separation as
\begin{equation}
d_{\ell}^{\mathrm{SU-SS}}
=
\mathbb{E}[\theta_\ell(S,U)]
-
\mathbb{E}[\theta_\ell(S,S)],
\end{equation}
and
\begin{equation}
d_{\ell}^{\mathrm{SU-UU}}
=
\mathbb{E}[\theta_\ell(S,U)]
-
\mathbb{E}[\theta_\ell(U,U)].
\end{equation}

The estimates are averaged across $r$ runs. Layers where S--U representations show consistently greater angular separation than both within-condition comparisons are treated as candidate safety-sensitive regions. The same divergence procedure is applied to Qwen3-VL-4B and InternVL2.5-4B.

\subsection{Activation Scaling}

To test sensitivity to activation magnitude, we use forward hooks to modify the outputs of selected Qwen3-VL-4B layers without changing model parameters. For a target activation tensor $\mathbf{h}$,
\begin{equation}
\mathbf{h}_{\mathrm{scaled}}=\alpha\mathbf{h},
\end{equation}
where
$\alpha\in\{0.85,0.9,1.0,1.15,1.2\}$.
Scaling is applied uniformly to all visual tokens in the selected layer region. We evaluate whether changes in activation magnitude alter refusal behavior as measured by URR and SRR.

\subsection{Directional Activation Steering}

To test whether safety-related representations define a controllable direction, we construct a layer-specific mean-difference direction
\begin{equation}
\mathbf{E}_{1,\ell}
=
\boldsymbol{\mu}_{\ell}^{\mathrm{unsafe}}
-
\boldsymbol{\mu}_{\ell}^{\mathrm{safe}},
\end{equation}
and normalize it as
\begin{equation}
\hat{\mathbf{E}}_{1,\ell}
=
\frac{\mathbf{E}_{1,\ell}}
{\|\mathbf{E}_{1,\ell}\|_2}.
\end{equation}

For token representation $\mathbf{h}_{\ell,t}$, its projection onto the learned direction is
\begin{equation}
\mathbf{P}_{\ell,t}
=
\left(
\mathbf{h}_{\ell,t}^{\top}
\hat{\mathbf{E}}_{1,\ell}
\right)
\hat{\mathbf{E}}_{1,\ell},
\end{equation}
and steering is applied as
\begin{equation}
\mathbf{h}'_{\ell,t}
=
\mathbf{h}_{\ell,t}
+
\alpha\mathbf{P}_{\ell,t},
\end{equation}
where
$\alpha\in\{0,0.5,1,2,4,-2,-4\}$.

We compare two intervention scopes: (1) focused steering applied to the CLS token and first 10 patch tokens, and (2) global steering applied to all visual tokens. This comparison tests whether behavioral control is concentrated in a small subset of visual representations or distributed across the visual token sequence.

\section{Results}

We evaluate our framework on Qwen3-VL-4B~\cite{Bai2023QwenVLAV} and use InternVL2.5-4B for cross-model validation of layer-wise activation geometry. Causal representation-level interventions are performed on Qwen3-VL-4B, while InternVL2.5-4B is used to assess whether the observed divergence patterns recur across architectures.

Our primary behavioral metrics are Unsafe Rejection Rate (URR) and Safe Rejection Rate (SRR), corresponding to rejection rates on unsafe and safe inputs, respectively. We use UnsafeBench~\cite{qu2025unsafebench} for harmful and ``hard'' safe examples, Google Conceptual Captions (GCC) for ``clean'' safe examples, and MMLU~\cite{hendrycks2020measuring} to assess general reasoning utility.

\subsection{Divergence Analysis}

We use the Mean Angular Difference defined in Section~4.1 to measure how strongly Safe--Unsafe (SU) representations diverge relative to Safe--Safe (SS) and Unsafe--Unsafe (UU) pairs across network depth.

\begin{figure}[t]
\centering
\begin{tabular}{cc}
\includegraphics[width=0.46\linewidth]{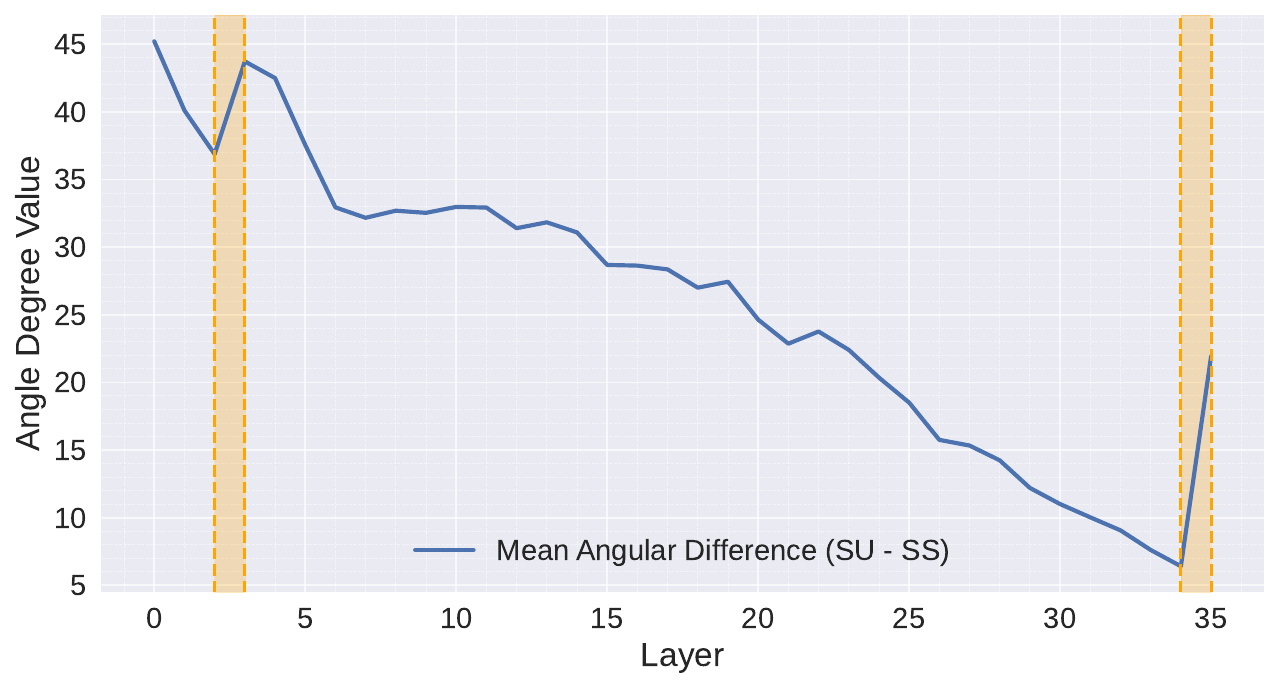} &
\includegraphics[width=0.46\linewidth]{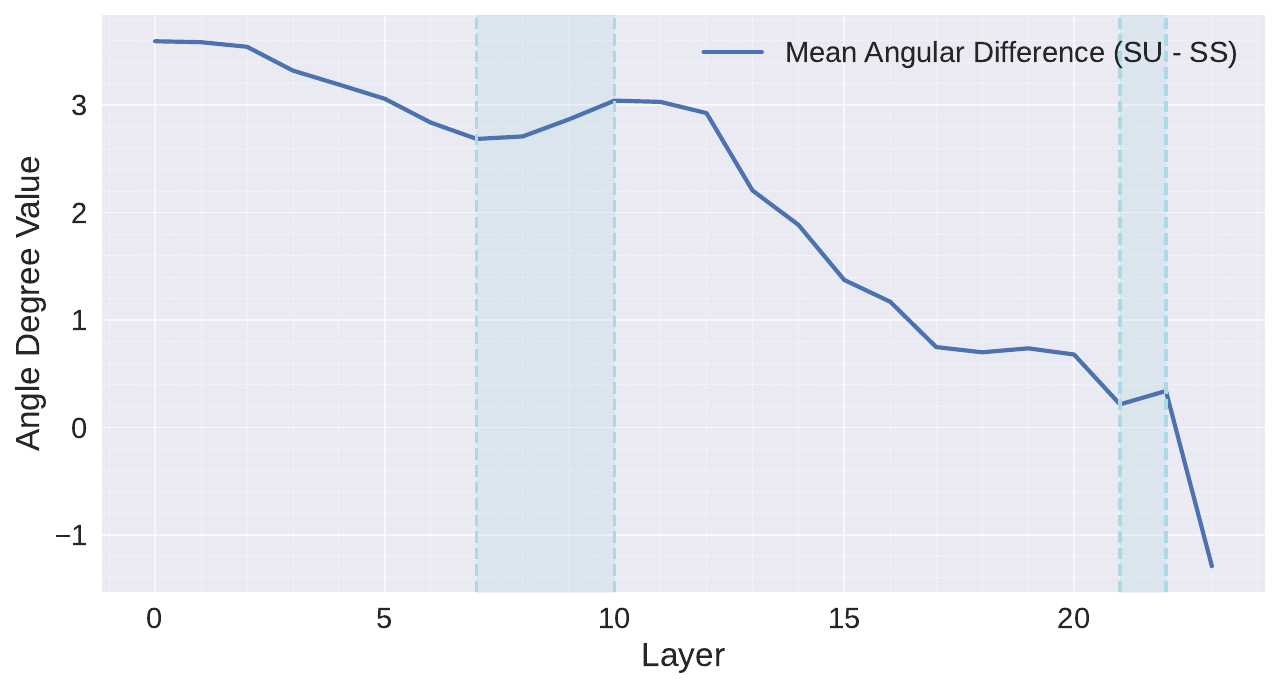} \\
{\small (a)} & {\small (b)} \\[6pt]

\includegraphics[width=0.46\linewidth]{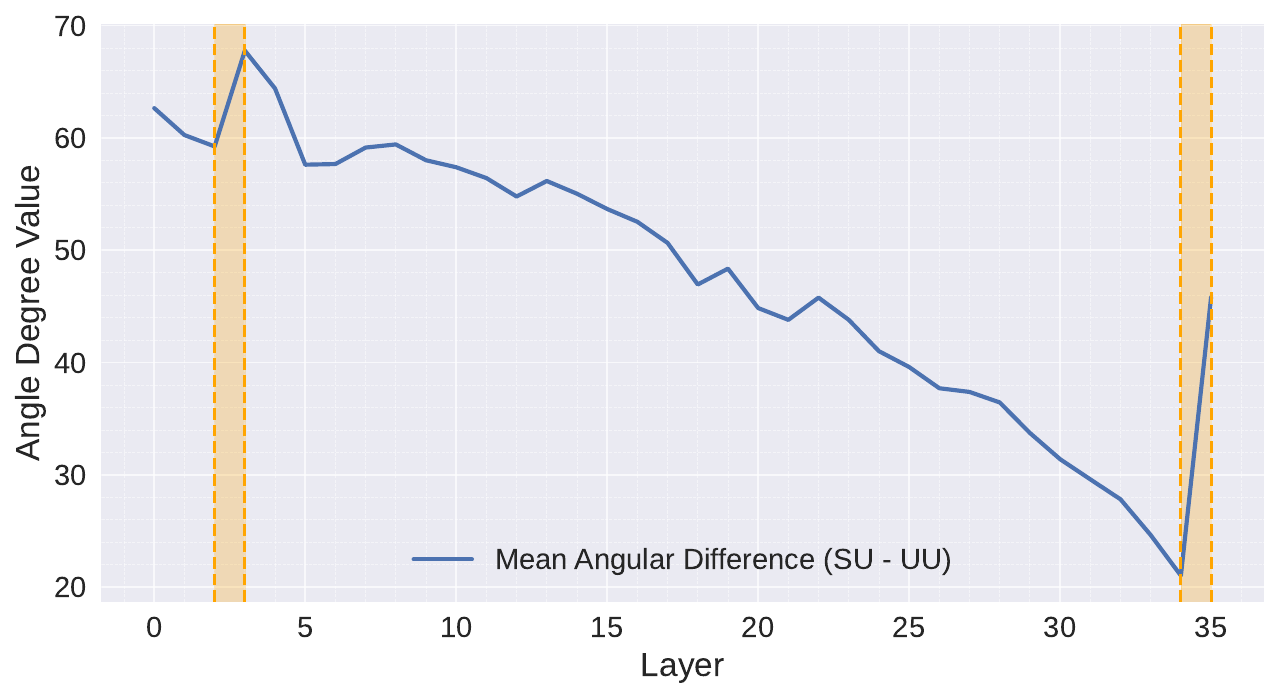} &
\includegraphics[width=0.46\linewidth]{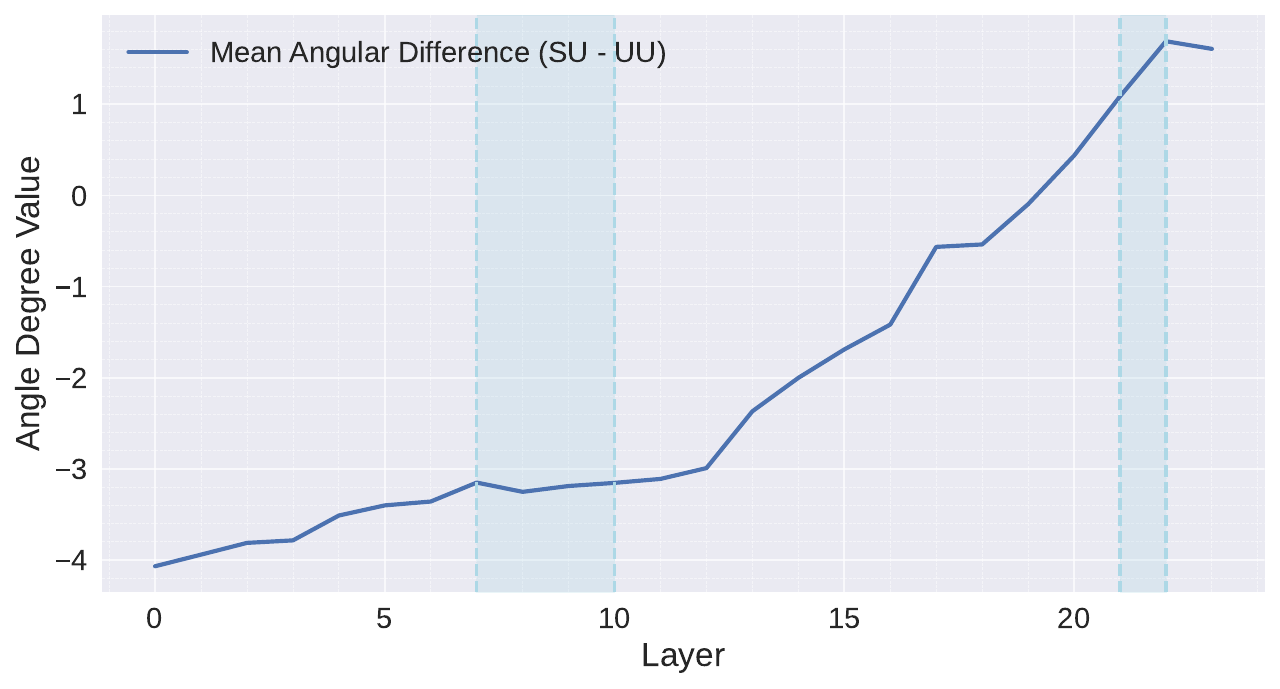} \\
{\small (c)} & {\small (d)}
\end{tabular}

\caption{
Layer-wise Mean Angular Divergence for Qwen3-VL-4B.
(a,c) The language tower exhibits strong early directional separation that generally decreases with depth, with SU--UU remaining larger than SU--SS across most layers.
(b,d) The vision encoder exhibits non-linear, heterogeneous divergence with distinct safety-sensitive regions around Layers 7--10 and 18--22.
}
\label{fig:angular_divergence_qwen}
\end{figure}

\paragraph{Qwen3-VL-4B.}
Figure~\ref{fig:angular_divergence_qwen} shows distinct depth-dependent patterns across the language and vision components. In the language tower, both SU--SS and SU--UU divergence are strongest in early layers and generally decrease with depth, with SU--UU remaining larger across most of the network.

\begin{figure}[t]
\centering
\includegraphics[width=0.9\linewidth]{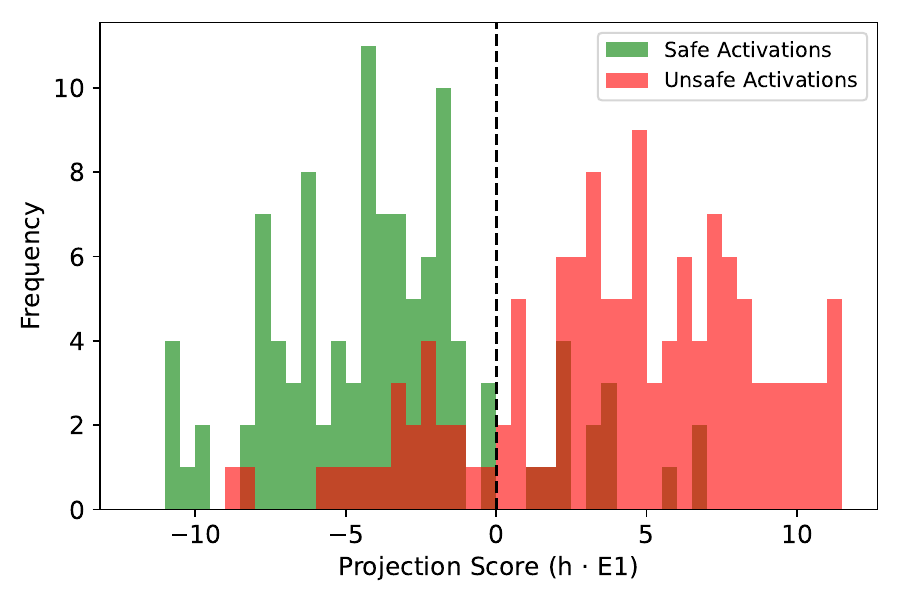}
\caption{
Projection of CLS token activations onto the learned safety direction ($\mathbf{E}_1$) at Layer 20 of Qwen3-VL-4B. Safe and unsafe inputs form clearly shifted distributions with limited overlap, indicating strong representational separability along a safety-related direction in deeper vision layers. Subsequent intervention results show that such separability does not necessarily imply equivalent directional controllability.
}
\label{fig:cav_projection_late}
\end{figure}

The vision encoder exhibits more heterogeneous behavior, with distinct regions around Layers 7--10 and 18--22. We select these regions for subsequent interventions to test whether their representational differences correspond to causal influence on refusal behavior. At Layer 20, projecting CLS activations onto the learned safety direction $\mathbf{E}_1$ produces clear separation between safe and unsafe examples (Figure~\ref{fig:cav_projection_late}), indicating strong representational separability at depth. As shown later, however, this separation does not necessarily imply equivalent directional controllability.

\begin{figure}[t]
\centering
\begin{tabular}{cc}
\includegraphics[width=0.46\linewidth]{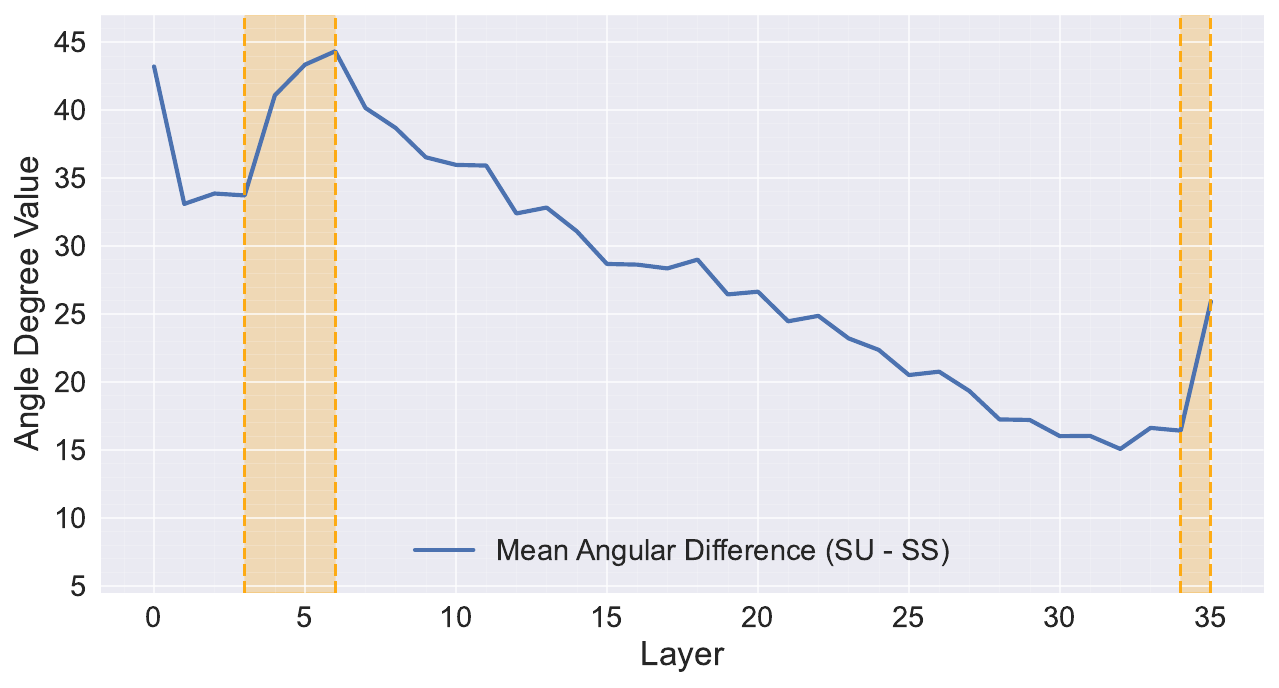} &
\includegraphics[width=0.46\linewidth]{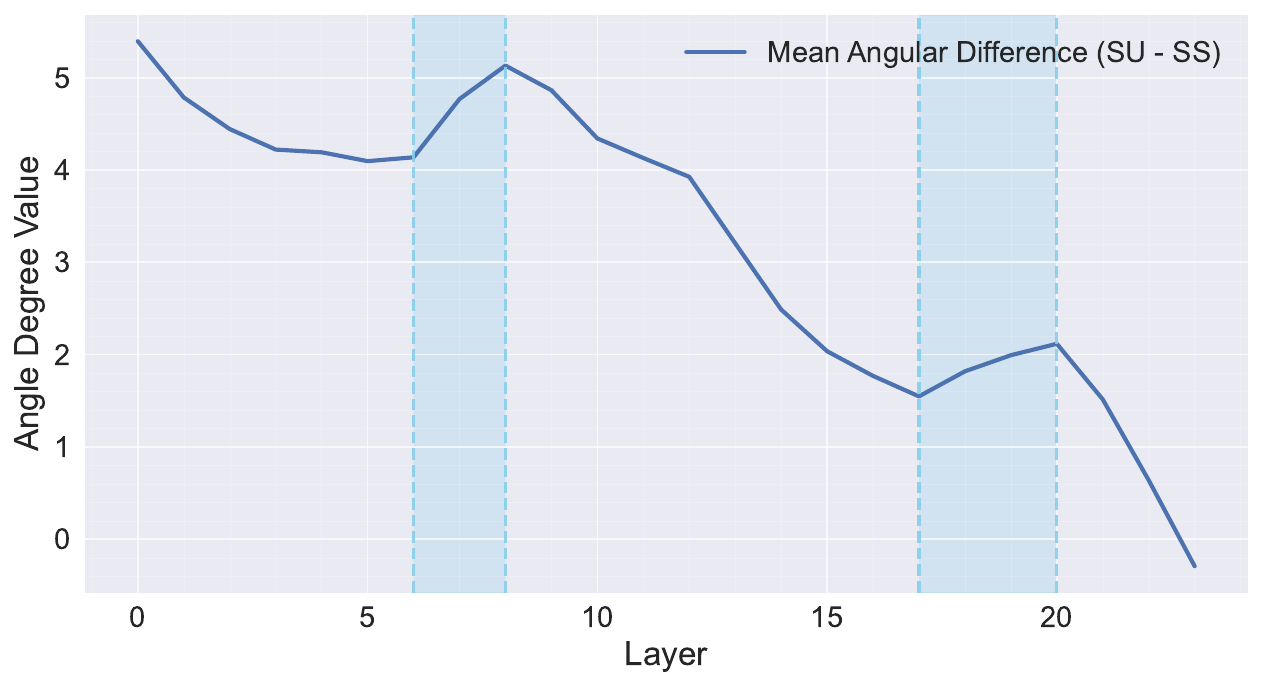} \\
{\small (a)} & {\small (b)} \\[6pt]

\includegraphics[width=0.46\linewidth]{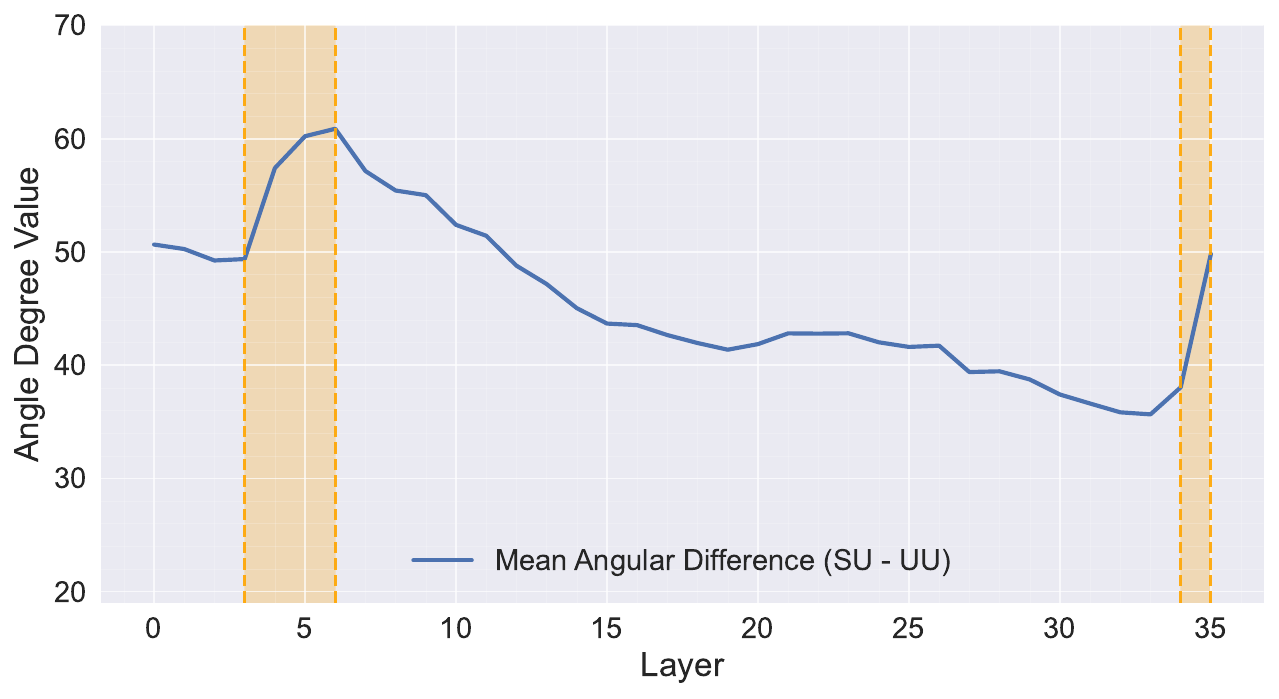} &
\includegraphics[width=0.46\linewidth]{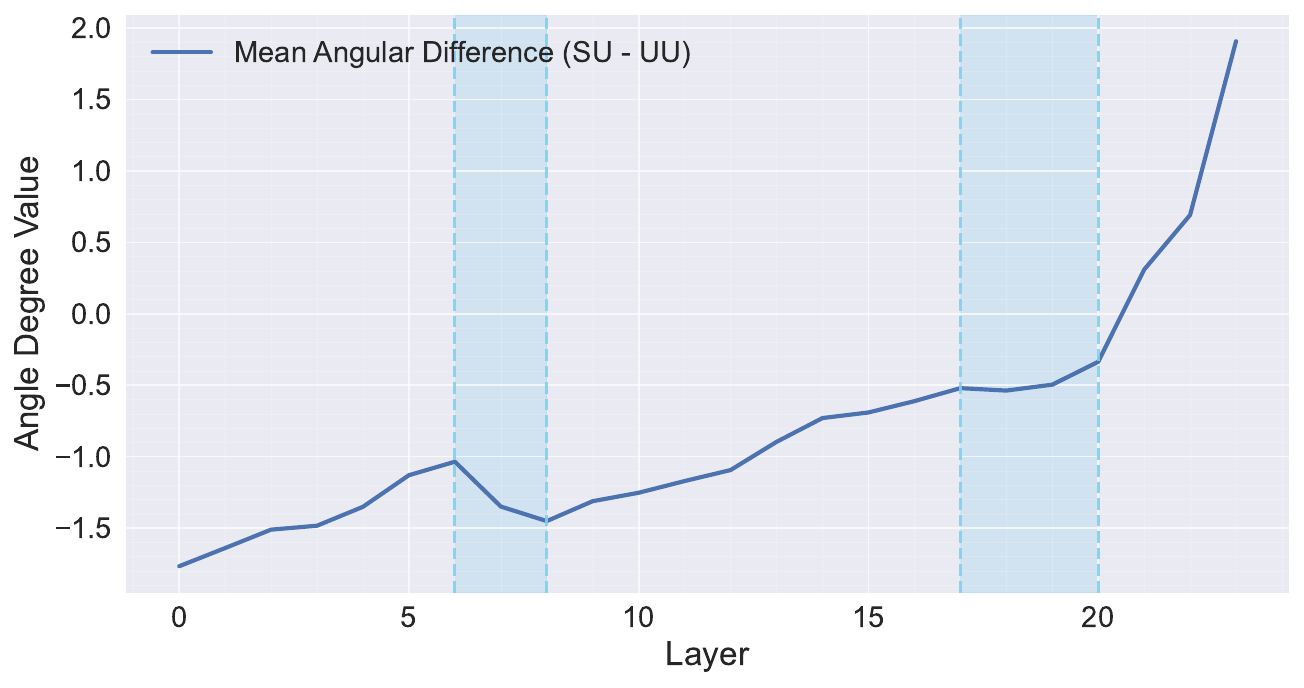} \\
{\small (c)} & {\small (d)}
\end{tabular}

\caption{
Layer-wise Mean Angular Divergence for InternVL2.5-4B.
(a,c) The language tower shows stronger early angular divergence that generally decreases with depth, qualitatively matching Qwen3-VL-4B.
(b) Vision SU--SS divergence is stronger in early and intermediate layers and decreases toward later depth.
(d) Vision SU--UU transitions from negative to positive divergence across depth, again qualitatively resembling Qwen3-VL-4B.
The precise layer locations and divergence magnitudes differ across architectures, suggesting recurring depth-dependent safety geometry with model-dependent localization.
}
\label{fig:angular_divergence_internvl}
\end{figure}

\paragraph{Cross-Model Validation with InternVL2.5-4B.}
Repeating the same analysis on InternVL2.5-4B reveals qualitatively similar depth-dependent organization (Figure~\ref{fig:angular_divergence_internvl}). Its language tower likewise exhibits stronger SU--SS and SU--UU divergence in earlier layers followed by a general decline with depth, although the strongest separation occurs at different absolute layer indices.

The vision encoder also shows non-uniform geometry: SU--SS divergence is stronger in early and intermediate representations and weakens toward later layers, while SU--UU transitions from negative toward positive divergence across depth, qualitatively resembling the trajectory observed in Qwen3-VL-4B.

Overall, both models exhibit systematic, non-uniform changes in safety-related representation geometry across depth, while differing in the precise layer locations and divergence magnitudes. These results provide cross-model representational evidence for qualitatively recurring depth-dependent safety geometry, but not causal equivalence; steering and scaling interventions are evaluated only on Qwen3-VL-4B.

\subsection{Parameter Scaling}

We evaluate sensitivity to activation magnitude using
$\alpha \in \{0.85,0.9,1.0,1.15,1.2\}$.
Results are shown in Tables~\ref{tab:param_scaling} and~\ref{tab:gcc_scaling}.

\begin{table}[t]
\centering
\caption{Parameter scaling results on Unsafe-bench (Qwen3-VL-4B). URR denotes Unsafe Rejection Rate and SRR denotes Safe Rejection Rate.}
\label{tab:param_scaling}

\begin{tabular}{ccccccc}
\toprule
\multirow{2}{*}{\textbf{$\alpha$}} &
\multicolumn{2}{c}{\textbf{Vision 7--10}} &
\multicolumn{2}{c}{\textbf{Vision 18--22}} &
\multicolumn{2}{c}{\textbf{Language}} \\
\cmidrule(lr){2-3}
\cmidrule(lr){4-5}
\cmidrule(lr){6-7}
& \textbf{URR} & \textbf{SRR}
& \textbf{URR} & \textbf{SRR}
& \textbf{URR} & \textbf{SRR} \\
\midrule
0.85 & 100 & 99 & 79 & 43 & 37 & 8 \\
0.90 & 95 & 89 & 94 & 72 & 80 & 49 \\
1.00 & 72 & 26 & 72 & 26 & 72 & 26 \\
1.15 & 72 & 23 & 96 & 91 & 97 & 77 \\
1.20 & 72 & 24 & 88 & 89 & 98 & 98 \\
\bottomrule
\end{tabular}
\end{table}

\begin{table}[t]
\centering
\caption{Parameter scaling results on Google Conceptual Captions (Qwen3-VL-4B). SRR denotes Safe Rejection Rate (\%).}
\label{tab:gcc_scaling}

\begin{tabular}{cccc}
\toprule
\textbf{$\alpha$} &
\textbf{Vision 7--10} &
\textbf{Vision 18--22} &
\textbf{Language} \\
\midrule
0.85 & 100 & 16 & 0 \\
0.90 & 100 & 16 & 0 \\
1.00 & 0   & 0  & 0 \\
1.15 & 9   & 8  & 6 \\
1.20 & 7   & 8  & 5 \\
\bottomrule
\end{tabular}
\end{table}

\noindent\textbf{Vision Layers 7--10:}
Early layers are highly sensitive to suppression: $\alpha=0.85$ yields 100\% URR on UnsafeBench but also 100\% SRR on GCC, indicating near-total refusal rather than selective safety improvement. In contrast, amplification produces little change from the 72\% baseline URR, revealing an asymmetric response to magnitude perturbation.

\noindent\textbf{Vision Layers 18--22:}
Later layers respond more strongly to amplification, with $\alpha=1.15$ increasing URR to 96\%. However, SRR simultaneously rises to 91\% on UnsafeBench while remaining 8--9\% on GCC, revealing a substantial safety--selectivity trade-off.

\noindent\textbf{Language Layers:}
The language tower also exhibits strong magnitude sensitivity: increasing $\alpha$ to 1.2 raises URR to 98\%, whereas reducing it to 0.85 lowers URR to 37\%.

\noindent\textbf{General Utility:}
On 100 MMLU samples, the tested scaling intervention produces no change from the 44\% baseline accuracy. However, the elevated SRR under several settings shows that unchanged MMLU performance does not imply preserved safety selectivity.

\subsection{Concept Activation Steering}

We steer the identified vision regions along $\mathbf{E}_1$, comparing focused interventions on the CLS and first 10 patch tokens with global interventions over all visual tokens (Table~\ref{tab:steering_results_combined}).

\begin{table*}[t]
\centering
\caption{Causal mediation analysis with activation steering on Qwen3-VL-4B. 
URR denotes Unsafe Rejection Rate and SRR denotes Safe Rejection Rate.}
\label{tab:steering_results_combined}

\begin{tabular}{c cc cc cc cc}
\toprule
& \multicolumn{4}{c}{\textbf{Layers 7--10}}
& \multicolumn{4}{c}{\textbf{Layers 18--20}} \\
\cmidrule(lr){2-5}
\cmidrule(lr){6-9}

\textbf{$\alpha$}
& \multicolumn{2}{c}{\textbf{Focused}}
& \multicolumn{2}{c}{\textbf{Global}}
& \multicolumn{2}{c}{\textbf{Focused}}
& \multicolumn{2}{c}{\textbf{Global}} \\
\cmidrule(lr){2-3}
\cmidrule(lr){4-5}
\cmidrule(lr){6-7}
\cmidrule(lr){8-9}

& \textbf{URR} & \textbf{SRR}
& \textbf{URR} & \textbf{SRR}
& \textbf{URR} & \textbf{SRR}
& \textbf{URR} & \textbf{SRR} \\
\midrule

Baseline ($\alpha=0$)
& 84 & 32
& 84 & 32
& 84 & 32
& 84 & 32 \\

0.5
& 84 & 32
& 84 & 24
& 82 & 26
& 82 & 28 \\

1.0
& 84 & 28
& \textbf{94} & \textbf{60}
& 82 & 30
& 82 & 32 \\

2.0
& 84 & 32
& 70 & 44
& 82 & 30
& 84 & 38 \\

4.0
& 82 & 30
& 6 & 8
& 82 & 30
& 80 & 66 \\

-2.0
& 82 & 30
& 82 & 30
& 82 & 26
& 82 & 26 \\

-4.0
& 82 & 30
& \textbf{96} & \textbf{88}
& 82 & 26
& 84 & 20 \\

\bottomrule
\end{tabular}

\vspace{0.4em}

\begin{minipage}{0.96\textwidth}
\small
\textbf{Notes:} URR = Unsafe Rejection Rate (true positive); 
SRR = Safe Rejection Rate (false positive). 
Focused steering is applied to CLS + 10 tokens, while global steering 
is applied to all tokens.
\end{minipage}

\end{table*}

\noindent\textbf{Layers 7--10:}
Global steering strongly modulates refusal behavior. Relative to the 84\% baseline URR, $\alpha=1.0$ increases URR to 94\%, while $\alpha=4.0$ reduces it to 6\%, substantially suppressing refusal. Steering in the opposite direction ($\alpha=-4.0$) raises URR to 96\%. Focused steering produces much smaller changes, suggesting that the causal effect is distributed across visual tokens rather than concentrated in the CLS representation.

\noindent\textbf{Layers 18--20:}
Later layers are comparatively insensitive to directional steering, with URR remaining near baseline across the tested strengths. This contrasts with their strong response to magnitude scaling. Despite clear geometric separation along $\mathbf{E}_1$ (Figure~\ref{fig:cav_projection_late}), steering this direction has limited behavioral effect, showing that representational separability does not necessarily imply causal controllability.

Overall, the interventions reveal a depth-dependent asymmetry: early vision layers permit strong directional control of refusal behavior, whereas later layers are more responsive to magnitude perturbations.

\section{Discussion}
\paragraph{Depth-Dependent Safety Mechanisms.}
Our results reveal distinct intervention profiles across the vision encoder of Qwen3-VL-4B. Early Layers 7--10 support strong bidirectional refusal control under global Concept Activation Steering, whereas later Layers 18--20 remain comparatively stable under the same directional intervention. Magnitude scaling affects both regions differently: early layers are highly sensitive to suppression, while later layers respond more strongly to amplification. These findings indicate a depth-dependent difference in intervention sensitivity rather than a strict directional-versus-magnitude dichotomy. Importantly, representational separability does not necessarily imply causal controllability: deeper activations remain well separated along $\mathbf{E}_1$, yet steering this direction produces limited behavioral change.
\paragraph{Cross-Model Safety Geometry.}
InternVL2.5-4B exhibits qualitatively similar depth-dependent safety geometry. In both models, language-layer divergence is stronger earlier and generally decreases with depth, while the vision encoders show non-uniform organization and systematic changes in SU--SS and SU--UU divergence. However, the precise magnitudes and layer locations differ across architectures. This suggests that safety-related geometry may recur across VLMs without implying a universal set of absolute ``safety layers.'' Because causal interventions are currently performed only on Qwen3-VL-4B, the InternVL2.5-4B results establish representational consistency rather than causal equivalence.
\paragraph{Security and Safety--Selectivity Implications.}
The bidirectional manipulation of refusal behavior exposes a white-box, representation-level attack surface. Early-layer steering can substantially strengthen or suppress refusal without modifying model parameters, identifying internal directions that input-level attacks could potentially target. However, our experiments do not demonstrate an end-to-end adversarial image or prompt that realizes the same shifts. The results also reveal a safety--selectivity trade-off: magnitude scaling can increase URR while simultaneously increasing SRR on difficult safe inputs. Although no change is detected in our 100-sample MMLU evaluation, this limited result does not establish preservation of broad utility.
\paragraph{Limitations and Future Work.}
Causal interventions are limited to Qwen3-VL-4B, while InternVL2.5-4B provides only qualitative representational validation. Future work should test equivalent interventions across architectures and quantify cross-model agreement. Our refusal-bypass experiments also require direct access to internal activations, leaving input-level realization as an open problem. Finally, broader evaluation across held-out splits, safety categories, and higher-dimensional safety subspaces is needed to assess direction stability and develop robust inference-time defenses.

\section{Conclusion}

We introduced a lightweight inference-time framework for identifying and manipulating safety-sensitive representations in Vision-Language Models without modifying model parameters. On Qwen3-VL-4B, selected early vision layers support strong bidirectional refusal control through Concept Activation Steering, while selected later layers are more responsive to activation magnitude scaling, reaching 96\% unsafe rejection at the cost of elevated false rejection. InternVL2.5-4B exhibits qualitatively similar depth-dependent activation geometry despite differences in precise layer location, providing cross-model representational evidence that this structure is not unique to Qwen3-VL-4B. Overall, our findings distinguish representational separability from causal controllability and expose a white-box representation-level attack surface that motivates inference-time stabilization of safety-sensitive representations.

\clearpage
\bibliographystyle{plainnat}
\bibliography{references}

\end{document}